\documentclass[10pt,twocolumn,letterpaper]{article}

\usepackage{cvpr}              % To produce the CAMERA-READY version
\definecolor{cvprblue}{rgb}{0.21,0.49,0.74}
\usepackage[pagebackref,breaklinks,colorlinks,citecolor=cvprblue]{hyperref}

\usepackage{adjustbox}
\usepackage{multirow}
\usepackage{colortbl}
\usepackage{xspace}
\usepackage[capitalize]{cleveref}
\usepackage{multirow}
\usepackage{float}
\usepackage{booktabs}
\usepackage[table, HTML]{xcolor}
\usepackage{bbding} 
\usepackage{pifont} 
\def\paperID{7505} % *** Enter the Paper ID here
\def\confName{CVPR}
\def\confYear{2026}

\usepackage[table]{xcolor}

\definecolor{resblue}{rgb}{0.64, 0.87, 0.93}
\definecolor{bestG}{rgb}{0.64, 0.71, 0.40}
\definecolor{resBlue}{HTML}{C2E2FA}
\definecolor{lBlue}{HTML}{D5ECFB}
\definecolor{lBlue1}{HTML}{EAF4FD}

\title{WPT: World-to-Policy Transfer via Online World Model Distillation}

\author{
    Guangfeng Jiang\textsuperscript{\rm 1},
    Yueru Luo\textsuperscript{\rm 2},
    Jun Liu\textsuperscript{\rm 1}$\textsuperscript{\ding{41}}$,
    Yi Huang\textsuperscript{\rm 2},
    Yiyao Zhu\textsuperscript{\rm 3},  \\
    Zhan Qu\textsuperscript{\rm 4},
    Dave Zhenyu Chen\textsuperscript{\rm 4},
    Bingbing Liu\textsuperscript{\rm 4},
    Xu Yan\textsuperscript{\rm 4}$\textsuperscript{\ding{41}}$\\
    \textsuperscript{1}University of Science and Technology of China, \\
    \textsuperscript{2}CUHK-SZ,
    \textsuperscript{3}HKUST,
    \textsuperscript{4}Huawei Foundation Model Department
}

\begin{document}
\maketitle
\begin{abstract}
Recent years have witnessed remarkable progress in world models, which primarily aim to capture the spatio-temporal correlations between an agent’s actions and the evolving environment.
However, existing approaches often suffer from tight runtime coupling or depend on offline reward signals, resulting in substantial inference overhead or hindering end-to-end optimization.
To overcome these limitations, we introduce \textbf{WPT}, a World-to-Policy Transfer training paradigm that enables online distillation under the guidance of an end-to-end world model.
Specifically, we develop a trainable reward model that infuses world knowledge into a teacher policy by aligning candidate trajectories with the future dynamics predicted by the world model.
Subsequently, we propose policy distillation and world reward distillation to transfer the teacher’s reasoning ability into a lightweight student policy, enhancing planning performance while preserving real-time deployability.
Extensive experiments on both open-loop and closed-loop benchmarks show that our WPT achieves state-of-the-art performance with a simple policy architecture: it attains a \textbf{0.11 collision rate} (open-loop) and achieves a \textbf{79.23 driving score} (closed-loop), surpassing both world-model-based and imitation-learning methods in accuracy and safety. Moreover, the student sustains up to \textbf{4.9$\times$ faster} inference, while retaining most of the gains.
\end{abstract}

% 最后加一个指标详细的，加速多少， intro也加    
% \footnote{$^*$ Work done during an internship at Huawei.}
\footnote{$\textsuperscript{\ding{41}}$ Corresponding authors.}
\section{Introduction}
\label{sec:intro}
Recent advances in autonomous driving have increasingly centered on world models that learn to capture the spatiotemporal dynamics of complex driving environments~\cite{Wang2023drive-wm, Yang2024driveoccworld, Li2025wote, Shang2025DriveDPO, Zheng2025World4Drive, zhang2025epona}.
Unlike imitation-based frameworks~\cite{Hu2022uniad, Jiang2023vad, Hu2022STP3, hu2025vision} that focus on replicating expert behavior, world-model-based methods explicitly model the causal interactions between agents and their surroundings, enabling anticipatory reasoning and more reliable long-horizon planning.

\begin{figure}[t]
    \centering
    \includegraphics[width=0.9\linewidth]{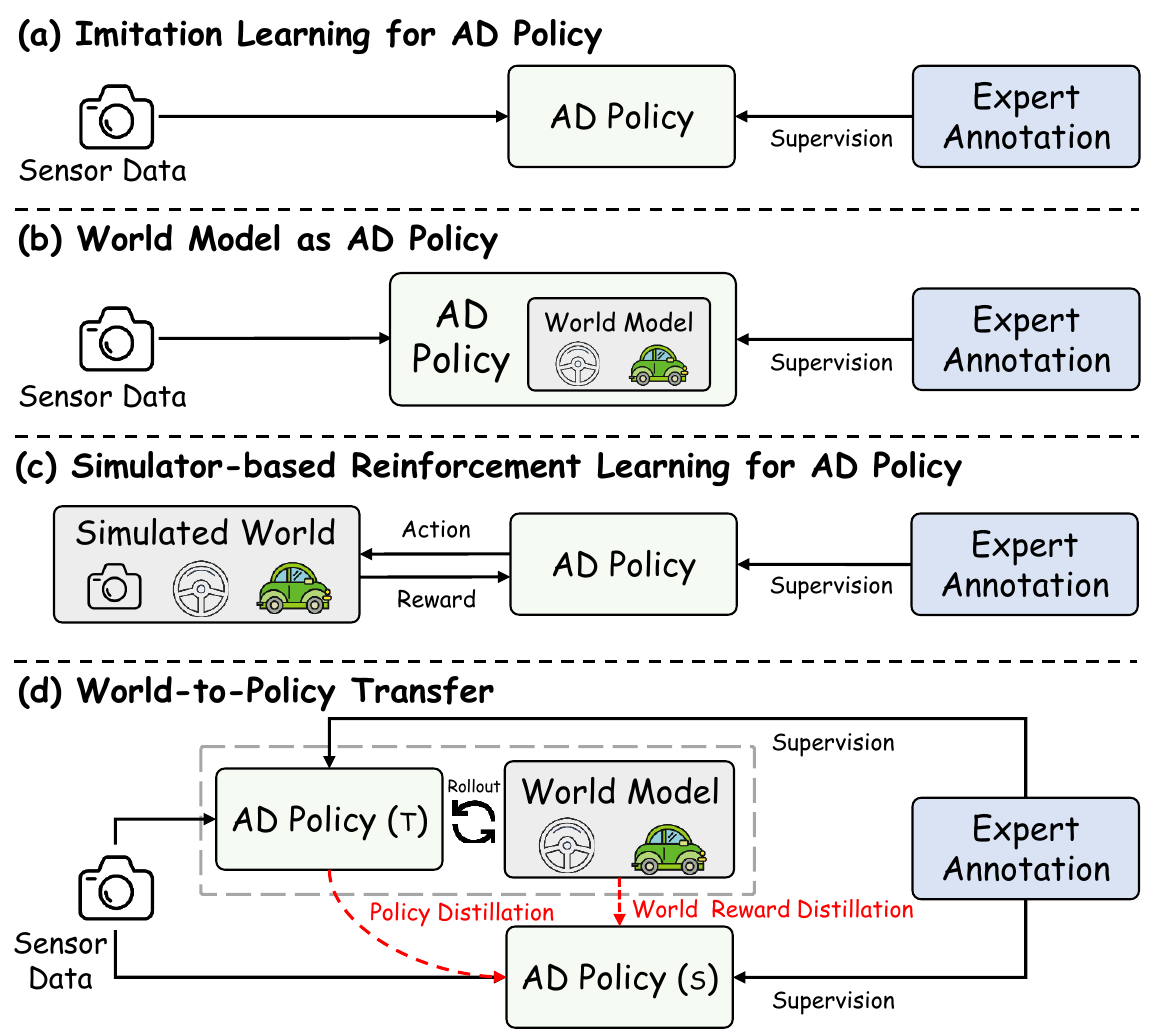}
    % \vspace{-2mm}
    \caption{\textbf{Different training paradigms of AD policy with world model.}
    (a) Imitation learning where the policy is trained using expert supervision. 
    (b) World model (WM) directly integrated into the AD policy for enhanced feature evolution and trajectory reasoning. 
    (c) Simulator-based reinforcement learning for AD policy training using a simulated world. 
    (d) Our WPT, where the policy interacts with the WM during training, with both the teacher policy (\textit{T}) and the student policy (\textit{S}) leveraging the WM for knowledge transfer. After training, the WM will be discarded.
    }
    % \vspace{-6mm}
    \label{fig:intro_compare}
\end{figure}

A world model (WM) generally predicts future scenarios from past observations and was originally introduced for simulated control and robotic applications~\cite{Schmidhuber2015Onlearningtothink, Ha2018WorldModels, Hafner2020MasteringAW, Wu2022DayDreamer, Hafner2023MasteringDD, Dawid2023IntroductionTL, Hafner2019Dreamtocontrol}.
Based on their role in enhancing autonomous driving (AD) policies (Fig.~\ref{fig:intro_compare}(a)), world models can be broadly categorized into two types:
The first category of methods directly integrates the world model into the driving policy (Fig.~\ref{fig:intro_compare}(b)), enabling more powerful feature evolution~\cite{zhang2025epona, Yang2024driveoccworld} and trajectory reasoning~\cite{Wang2023drive-wm, Li2025wote, Shang2025DriveDPO, Zheng2025World4Drive}.
Specifically, Drive-OccWorld~\cite{Yang2024driveoccworld} jointly models occupancy and flow prediction to support safe and interpretable trajectory planning. 
Alternatively, methods such as WoTE~\cite{Li2025wote} and DriveDPO~\cite{Shang2025DriveDPO} integrate an implicit bird-eye-view (BEV) world model with a learned reward evaluator to assess multiple trajectory candidates according to their predicted future states.
Despite their performance gains, these methods exhibit a strong dependence on accurate future predictions and autoregressive rollouts, where sequential dependencies between prediction steps substantially hinder real-time efficiency.
The second category of methods treats the world model as a simulator to enable closed-loop reinforcement learning for AD policy training~\cite{Gao2025RAD, Zhang2021EndtoEndUD, Lu2022ImitationIN, Toromanoff2019EndtoEndMR, dosovitskiy2017carla, Chen2021LearningTD}, as shown in Fig.~\ref{fig:intro_compare}(c).
Nevertheless, such approaches are highly dependent on the fidelity of the simulator’s generated data and are evaluated primarily in synthetic environments.

A natural question arises: \textit{how can the AD policy leverage the world knowledge while avoiding extra computational overhead?}
To this end, we present \textbf{WPT} (World-to-Policy Transfer in Fig.~\ref{fig:intro_compare}(d)), a novel training paradigm where the policy interacts with a world model during training to acquire predictive awareness of future dynamics, while maintaining real-time efficiency through a simple policy network during deployment.
Concretely, for a given AD policy, we utilize a world model to capture the spatiotemporal evolution of the environment from its learned representations.
We then design a trainable, interaction-based reward model that evaluates each candidate trajectory according to its consistency with the predicted future world states, thereby enabling the selection of the optimal trajectory.
The proposed interaction mechanism allows the end-to-end model to anticipate future environmental dynamics, enabling the policy to internalize world-model knowledge through predicted evolutions.
Furthermore, to satisfy real-time requirements, we introduce policy distillation and world reward distillation to transfer the reasoning capability of the large world model to a lightweight policy network, enabling fast inference while enhancing performance through knowledge transfer.

Overall, our \textbf{WPT} framework offers three key advantages:
% \textbf{1) Interactivity.} It establishes online trajectory–world interaction, enabling real-time and learnable decision-making by explicitly coupling the planner with the world model.
\textbf{(1) Interpretability.} It achieves end-to-end optimization through a reward-guided mechanism that bridges prediction and planning, resulting in world-consistent and explainable driving behaviors.
\textbf{(2) Efficiency.} It preserves real-time performance through distillation, transferring the reasoning ability of the large world model to a lightweight planner for fast inference (up to \textbf{4.9$\times$} speedup).
\textbf{(3) Effectiveness.} Experiments across both open-loop and closed-loop benchmarks show that WPT delivers state-of-the-art results (open-loop: \textbf{0.11\%} collision, closed-loop: \textbf{79.23} driving score) on different lightweight policies.

The main contributions are summarized as follows:
\begin{itemize}
    \item We propose \textbf{WPT}, an online distillation paradigm that integrates world modeling with a trainable reward model for interpretable and world-consistent planning.
    \item We design policy distillation and world reward distillation, transferring reasoning capabilities from a large model to a lightweight one.
    \item Extensive experiments on diverse benchmarks and baselines demonstrate that WPT achieves state-of-the-art performance, with {0.61m} L2 / {0.11\%} collision in open-loop and {79.23} driving score in closed-loop, outperforming existing world-model-based and imitation-learning-based methods. Moreover, these gains transfer to our student model with {up to 4.9$\times$} faster inference.
\end{itemize}
\begin{figure*}[ht]
    \centering
    \includegraphics[width=1.0\linewidth]{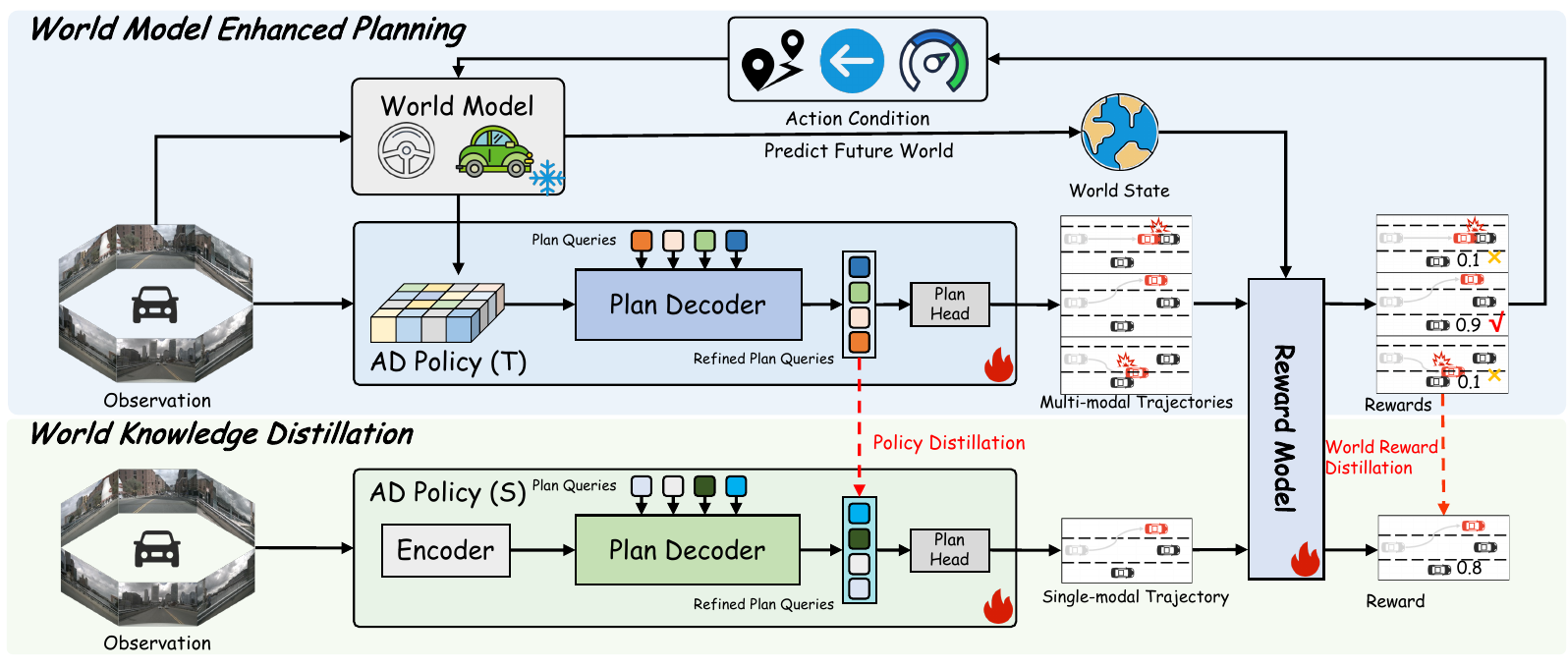}
    % \vspace{-6mm}
    \caption{\textbf{Overview of WPT framework.} During training (top), the pretrained world model predicts future world under given action conditions, and the teacher AD policy (T) generates multi-modal trajectories. The \emph{reward model} evaluates these trajectories to produce world reward. During distillation (bottom), the student AD policy (S) learns from the teacher through two mechanisms: (1) \emph{policy distillation}, which aligns the planning representations between teacher and student; and (2) \emph{world reward distillation}, which encourages the student to match the teacher’s optimal reward trajectory in the predicted future world.}
    % \vspace{-4mm}
    \label{fig:wpt}
\end{figure*}

\section{Related Work}
\subsection{End-to-End Autonomous Driving}
End-to-end autonomous driving directly maps raw multi-sensor inputs to future trajectories or low-level control commands \cite{2016end}. From the perspective of output modalities, recent research can be broadly categorized into single-modal and multi-modal trajectory planning approaches.

Single‑modal planning predicts one ``best" trajectory, as seen in approaches like UniAD~\cite{Hu2022uniad}, which integrates multi‑task modules in a unified framework optimized for planning. Further, VAD~\cite{Jiang2023vad} replaces dense scene representations with vectorized ones. 
Subsequent works refine this line along several axes:
\eg, sparse representation to improve efficiency~\cite{2025sparsedrive,zhang2024sparsead,2025drivetransformer,guo2025ipad},
temporal modeling to stabilize trajectory outputs~\cite{song2025don,Zhang2025bridgead}, and handling of online map uncertainty to increase planning reliability~\cite{gu2024uncad,yang2025uncad}.

To handle multi-future ambiguities, recent works explore multi-modal planning.
VADv2~\cite{chen2024vadv2} builds an anchor vocabulary for action spaces together with probabilistic planning to capture diverse future options. Hydra-MDP~\cite{li2024hydra,li2025hydra} advances this via multi‑expert distillation into a multi‑head student, yielding trajectories aligned with distinct criteria. 
More recently, diffusion‑based planners~\cite{Shang2025DriveDPO, zheng2025diffusion, xing2025goalflow} have emerged as a strong paradigm for modeling multi‑modal trajectory distributions, providing diversity naturally. Alongside generative modeling, the community has explored preference‑ or reward‑driven selection to better align outputs with different objectives: safety‑targeted selection~\cite{Shang2025DriveDPO}, human‑style alignment~\cite{Li2025FinetuningGT}, \etc.

\subsection{World Models for Driving}
The world models aim to learn a compact representation of the environment and predict future states based on an agent’s actions and past observations \cite{Wang2023drive-wm,lu20254d,Yang2024GeneralizedPM,Gao2024VistaAG,Wang2023DriveDreamerTR,liang2025worldlens}.

\noindent\textbf{Video World Model.} Video world models generate future visual frames or videos conditioned on candidate actions or trajectories, enabling planners to ``see'' what would happen under each choice and score options with perceptual (image-level) rewards.
Particularly, Drive-WM~\cite{Wang2023drive-wm} controllably generates multi-view videos under different maneuvers, and then selects trajectories via image-level rewards, showing the feasibility of WM-guided planning.
% DriveDreamer~\cite{Wang2023DriveDreamerTR} incorporates diverse driving conditions for real-world driving videos generation.
% GenAD~\cite{Yang2024GeneralizedPM} scales action-conditioned video prediction for driving by training on a large web-sourced driving data.
% Vista~\cite{Gao2024VistaAG} targets high-fidelity and long-horizon rollouts with versatile action controllability, enabling reward estimation without ground-truth actions. 
Most recently, Epona~\cite{zhang2025epona} adopts an autoregressive diffusion WM with spatial–temporal factorization for long-horizon rollouts, and integrates trajectory prediction with video generation for end-to-end planning. 

\noindent\textbf{Occupancy World Model.} Occupancy world models evolve the 3D scene volumetric states over time, offering planner-friendly rollouts in physical space. OccWorld~\cite{zheng2023occworld} tokenizes 3D occupancy and autoregressively predicts future occupancy and ego motion, enabling fine-grained scene evolution without dense labels. Drive-OccWorld~\cite{Yang2024driveoccworld} forecasts 4D occupancy and evaluates trajectories with occupancy costs.
RenderWorld~\cite{yan2025renderworld} encodes 3D occupancy into tokens, which a world model uses to forecast future 4D occupancy and ego motion. 
% GaussianAD~\cite{zheng2024gaussianad} proposes sparse 3D Gaussians as a compact substrate unifying perception, prediction, and planning.

\noindent\textbf{Latent World Model.} Latent world models roll out future latent features instead of pixels or voxels. {LAW}~\cite{li2024law} self-supervises an action-conditioned latent WM by predicting visual latent features from current features and trajectories. WoTE~\cite{Li2025wote} performs online trajectory evaluation using a BEV WM that forecasts future BEV states for multiple trajectory candidates. Their BEV WM predictions further are supervised via a traffic simulator, which also enables evaluating these trajectories.
% World4Drive~\cite{Zheng2025World4Drive} enriches a latent WM with intention cues and vision foundation model priors, generating multi-modal plans. Then, a WM selector commits to the optimal trajectory based on the predicted latent state.

Unlike these approaches, which rely on rollouts or online evaluation during deployment, WPT is a purely training-time paradigm. It provides interaction-based rewards and facilitates world knowledge transfer, enabling the deployed policy to plan in real time without additional overhead.

% 总结上述工作存在的问题，我们的方法优势

\section{Method}
\label{sec:method}

This section outlines the four core components of our proposed WPT framework: (1) the Autonomous Driving policy (Sec.~\ref{sec:e2e}), (2) the World Model (Sec.~\ref{sec:world_model}), (3) the Reward Model (Sec.~\ref{sec:reward_model}), and (4) the World Knowledge Distillation (Sec.~\ref{sec:distillation}).
As shown in Fig.~\ref{fig:wpt}, WPT leverages the world model during training to guide the teacher AD policy in generating future-aware trajectories. The student policy then distills knowledge from both the teacher’s planning representations and the world model’s reward supervision.

\subsection{Autonomous Driving Policy}
\label{sec:e2e}
This paper focuses on improving the planning performance of AD policies, employing a standard end-to-end architecture as depicted in~\cref{fig:wpt}.
The model takes as input a temporal sequence of multi-view camera images and encodes them into a unified world representation $F^{\text{w}}$  (\eg, BEV features) that captures the spatial context of the driving environment. 

Subsequently, a planning decoder $\mathcal{P}_D$ receives both the world representation $F^{\text{w}}$ and a set of learnable planning queries $Q$, and refines the queries through cross-attention interaction:
\begin{equation}
\small
\label{eq:plandecoder}
     \tilde{Q} =\mathcal{P}_D\left( Q,F^{\text{w}} \right),
\end{equation}
where $\mathcal{P}_D$ denotes a \textit{CrossAttention}-based module that facilitates interaction between planning queries and spatial scene representations.

Finally, a lightweight MLP-based plan head $\mathcal{P}_h$ decodes the refined queries into predicted future trajectories $\hat{\mathcal{T}}$:
\begin{equation}
\small
\label{eq:planhead}
     \hat{\mathcal{T}} =\mathcal{P}_h\left( \tilde{Q} \right).
\end{equation}

To support both performance and real-time deployment, we design two types of policies as shown in~\cref{fig:wpt}.

% a multi-modal AD policy and a single-modal AD policy.
\noindent\textbf{Multi-modal Policy as a Teacher.}
The multi-modal policy first generates a set of candidate trajectories 
$\mathcal{T} = \{ \tau_1, \ldots, \tau_N \}$ through the planning decoder, where $N$ is the number of candidate trajectories \cite{Hu2022uniad}.
These trajectories are evaluated through interaction with the world model to select the optimal trajectory under predicted future states.

\noindent\textbf{Single-modal Policy as a Student.}
To support real-time inference, we introduce a lightweight policy that predicts the entire future trajectory $\hat{\mathcal{T}}$ in a single forward process. 
% For each frame, the plan queries $Q^S \in \mathbb{R}^{n \times d}$ are initialized, where $n$ is the number of predicted trajectory points and $d$ is the query dimension. The trajectory is decoded via:
For each frame, the plan queries $Q^S$ are initialized. The trajectory is decoded via:
\begin{equation}
\small
\label{eq:student_planhead}
    \hat{\mathcal{T}} = \mathcal{P}_h\left(\mathcal{P}_D\left(Q^S, F^{\text{w}}\right)\right).
\end{equation}

Unlike the teacher policy, the student policy does not rely on multi-modal trajectory generation or world model interaction during inference. Instead, it directly extracts planning-relevant cues from world representation, enabling low-latency trajectory prediction.

% The first part of our design is an autoregressive planning network that generates multi-modal trajectories. We follow a sampling strategy similar to that in UniAD, where we generate a set of candidate trajectories \( T = \{ t_1, \ldots, t_n \} \) for the teacher model.

% Next, based on the predicted occupancy states from the world model, we select the top-\(k\) trajectories \( t_{\text{best}} \), following a cost function inspired by Drive-OccWorld. These selected trajectories are then decoded through a PlanDecoder, which is based on a cross-attention mechanism. The PlanDecoder generates the next frame’s trajectory \( \tilde{\tau} = \mathcal{P}_D(q_{\tau}, \tau^*, s) \), where \( q_{\tau} \) represents the query for the current time step, \( \tau^* \) is the top-\(k\) selected trajectory, and \( s \) is the current world state. This autoregressive approach leverages the world model’s ability to generate high-quality trajectories, resulting in a higher-performing autonomous driving policy.

\subsection{World Model}
\label{sec:world_model}

\textbf{Preliminaries.} 
Autonomous driving world models $\mathcal{W}$ are generative models that predict future driving environment states $s$ conditioned on observational data $o$ and action condition $c$ (\eg, historical trajectories, navigation commands, and ego-vehicle states):
\begin{equation}
\small
\label{eq:wm}
    \mathcal{W}\left( \left\{ o_{t-h},...,o_t \right\} ,\left\{ c_{t-h},...,c_t \right\} \right) =s_{t+1},
\end{equation}
where $h$ denotes the number of historical observations, $t$ is the current time. Subsequently, $s$ can be decoded into the corresponding world representation through the decoder of the corresponding mode, such as images, occupancy, \etc.

Next, we describe a general world model structure, which consists of three main components: observation encoder, feature aggregation, and world decoder.

\noindent\textbf{(1) Observation Encoder $\mathcal{W}_E$.} 
This component processes historical observations $\left\{ o_{t-h},...,o_t \right\}$ to extract features, which are then transformed into a world embedding $F^{\text{w}}_t$:
\begin{equation}
\small
\label{eq:history_encoder}
    F^{\text{w}}_t = \mathcal{W}_E\left( \left\{ o_{t-h}, ..., o_t \right\} \right).
\end{equation}

\noindent\textbf{(2) Feature Aggregation $\mathcal{A}_M$.} 
The feature aggregation module aggregates historical world embeddings $F^{\text{w}}_{t-h:t}$ to capture temporal context and ensure consistency across the historical sequence. Then, these features yield enhanced world representations $\tilde{F}^{\text{w}}_t$ for future prediction:
\begin{equation}
\small
\label{eq:memory_queue}
    \tilde{F}^{\text{w}}_t = \mathcal{A}_M\left( F^{\text{w}}_{t-h:t} \right).
\end{equation}

\noindent\textbf{(3) World Decoder $\mathcal{W}_D$.} 
The world decoder $\mathcal{W}_D$ is an autoregressive method that predicts the future world embedding $F^{\text{w}}_{t+1}$ based on action conditions $c$ (e.g., driving commands and historical trajectories) and the enhanced world features $\tilde{F}^{\text{w}}_t$ from $\mathcal{A}_M$. This process is formulated as:
\begin{equation}
\small
\label{eq:world_decoder}
    F^{\text{w}}_{t+1} = \mathcal{W}_D\left( \tilde{F}^{\text{w}}_t, c_t \right).
\end{equation}

% \noindent\textbf{Enhanced Planning.}
% % In our framework, we directly use the pretrained world model without updating its parameters. 
% To exploit the model's ability to anticipate future scene evolution, the multi-modal AD policy sets the world embedding $F^{\text{w}}$ of its $\mathcal{P}_D$ as predicted world state $F^{\text{w}}_{t+1}$ from the $\mathcal{W}_D$. 
% The planning formulation Eq.~(\ref{eq:plandecoder}) and Eq.~(\ref{eq:planhead}) are become:
% \begin{equation}
% \small
% \label{eq:wm_plan}
%     \mathcal{T}_{t+1} = \mathcal{P}_h\left(\mathcal{P}_D\left(Q^T, F^{\text{w}}_{t+1}\right)\right).
% \end{equation}

% This approach leverages the world model's predictive capability to refine the AD policy's planning, providing more accurate trajectory predictions while accounting for future environmental dynamics.

% 增加一个世界模型的总结，也就是observation和action(traj)的关系，世界模型的能力可以增强ad policy，然后引出下一章节的reward model
% 这种方法利用了世界模型的预测能力来
%~\cref{eq:plandecoder,eq:planhead} 
% Eqs. (~\ref{eq:plandecoder}) and (~\ref{eq:planhead})
\noindent\textbf{World Model Enhanced Planning.}
To exploit the model's ability to anticipate future scene evolution, the multi-modal AD policy sets the world embedding $F^{\text{w}}$ of its planning decoder $\mathcal{P}_D$ as the predicted world state $F^{\text{w}}_{t+1}$ from the world decoder $\mathcal{W}_D$. This autoregressive approach allows the planner to generate future trajectories that are aligned with the predicted world state, as depicted in~\cref{fig:wpt}. The planning formulation in~\cref{eq:plandecoder,eq:planhead} is then updated as:
\begin{equation}
\small
\label{eq:wm_plan}
    \mathcal{T}_{t+1} = \mathcal{P}_h\left(\mathcal{P}_D\left(Q^T, F^{\text{w}}_{t+1}\right)\right).
\end{equation}

This approach leverages the world model's predictive capability to refine the AD policy's planning, providing more accurate trajectory predictions while accounting for future environmental dynamics. However, while this method enhances the AD policy's capabilities, the predicted trajectories still primarily mimic expert behavior without fully accounting for the changing dynamics of the future world. To address this limitation, we develop a trainable reward model that transfers world knowledge into the teacher policy by aligning candidate trajectories with the future dynamics predicted by the world model. This allows for a more refined evaluation of the trajectories, enabling the policy to better adapt to the evolving environment, which we will discuss in the next section.

\begin{figure}[ht]
    \centering
     \includegraphics[width=8cm, height=10.5cm]{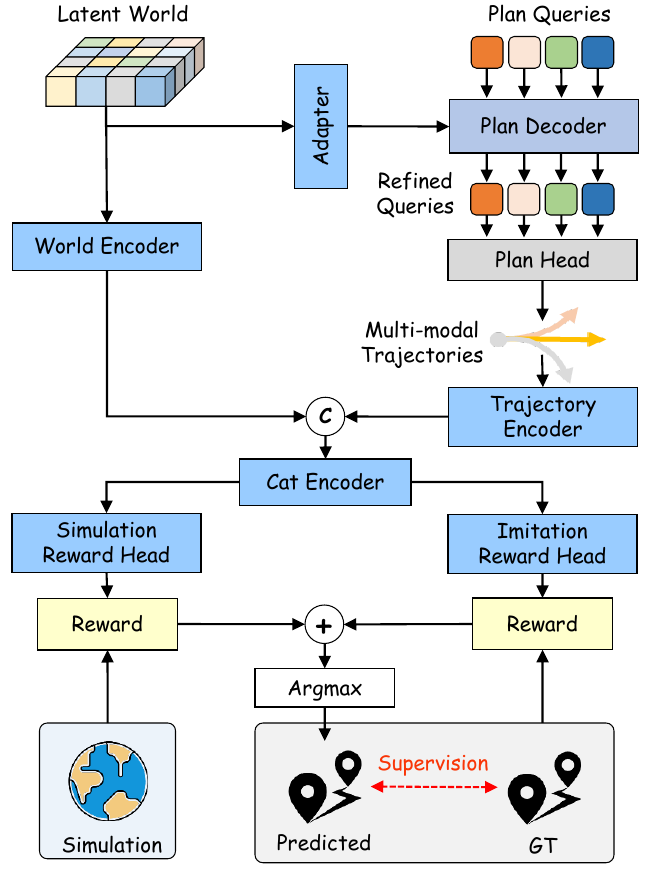}
    \caption{\textbf{Overview of reward model.} The reward model consists of multiple components: the world encoder processes the latent world representation, while the plan queries are refined through the plan decoder and plan head to generate multi-modal candidate trajectories. These trajectories are then passed to the trajectory encoder, which encodes them for evaluation by two distinct reward heads: the simulation reward head and the imitation reward head. The final reward is computed by combining these reward values, with the best trajectory selected via the argmax operation. The supervisory signals of the reward model come from simulation and imitation. For the detailed process, please refer to Sec. \ref{sec:reward_model}.}
    \label{fig:reward_model}
\end{figure}

\subsection{Reward Model}
\label{sec:reward_model}
In this section, we introduce a world model reward-based distillation mechanism that transfers the predictive capability of the world model into the AD policy. 
We design two complementary forms of reward supervision. 
The first is an imitation reward, where the reward model evaluates which trajectory aligns best with human driving preferences under future world evolution.
The second is a simulation reward, where the reward model assigns scores based on predicted world states and driving quality metrics, such as PDM scores in NAVISIM~\cite{dauner2024navsim}.

% \noindent\textbf{Reward Model.}
The reward model serves as the key mechanism for transferring the future predictive knowledge of the world model into the AD policy.
As shown in Fig.~\ref{fig:reward_model}, we evaluate each candidate trajectory $\tau_i$ by combining it with the predicted future world state $F^{\text{w}}_{t+1}$:
\begin{equation}
\small
    F_{w,i} = \text{RewardModel}\left(F^{\text{w}}_{t+1}, \tau_i\right),
\end{equation}
where $F_{w,i}$ denotes the joint trajectory–world interaction representation.
Two types of reward heads are then applied to obtain trajectory rewards, and the final best trajectory is:
\begin{equation}
\label{eq:argmax_traj}
\small
    \tau^* = \arg\max_{i} \left( w_1 r_{\text{im},i} + w_2 r_{\text{sim},i} \right),
\end{equation}
where $r_{\text{im}}$ and $r_{\text{sim}}$ represent the imitation and simulation reward respectively, $w_1$ and $w_2$ are balancing coefficients.

\noindent\textbf{Imitation Reward.}
The imitation reward aims to assess how well each candidate trajectory aligns with expert driving behavior. 
For each candidate trajectory $\tau_i$, we first compute its L2 distance $d_i$ from the corresponding expert trajectory, and then normalize it via a softmax function to obtain the target imitation score:
\begin{equation}
\small
    % r^{\text{im}*}_i = \text{Softmax}\left(-\frac{d_i}{\lambda}\right).
    r_{\text{im},i}^{*}=\text{softmax} \left(\frac{-d_i}{\sum_{j=1}^N{-d_j}}\right).
\end{equation}

The predicted imitation reward $r_{\text{im}}$ from the reward model is supervised by minimizing the cross-entropy loss with respect to the softmax-normalized target:
\begin{equation}
\small
    \mathcal{L}_{\text{im}} = \text{CrossEntropy}\left(r_{\text{im}}, r^{*}_{\text{im}}\right).
\end{equation}

This formulation encourages the reward model to assign higher scores to trajectories that closely match human driving patterns, thereby transferring human-like preferences to the AD policy.

\noindent\textbf{Simulation Reward.}
Unlike the imitation reward, which captures human driving preferences, the simulation reward evaluates candidate trajectories from an environment-centered perspective, emphasizing safety, comfort, and driving efficiency. 
Inspired by the NAVISIM~\cite{dauner2024navsim} score, we construct five metrics in the predicted future world: no collisions (NC), drivable area compliance (DAC), time-to-collision (TTC), ego progress (EP), and comfort (Comf). For implementation details, refer to appendix. The final simulation reward is expressed as:
\[
r_{\text{sim}}^* = \{ r_{\text{NC}}, r_{\text{DAC}}, r_{\text{TTC}}, r_{\text{EP}}, r_{\text{Comf}} \}.
\]
The predicted simulation reward $r_{\text{sim}}$ is supervised using binary cross entropy:
\begin{equation}
\small
    \mathcal{L}_{\text{sim}} = \text{BCE} \left( r_{\text{sim}}, r_{\text{sim}}^*\right).
\end{equation}

This supervision encourages the model to align its trajectory evaluation with safety and physical constraints derived from the predicted world.

\noindent\textbf{Final Reward.}
To balance human-like driving preference and environment-aware evaluation, the final reward integrates both imitation and simulation rewards as:
\begin{equation}
\small
\begin{aligned}
    r_{\text{final}} =\;&
    \alpha_1 \log r_{\text{im}}
    + \alpha_2 \log r_{\text{NC}}
    + \alpha_3 \log r_{\text{DAC}} \\
    &+ \alpha_4 \log \left( 5\, r_{\text{TTC}} + 5\, r_{\text{EP}} + 2\, r_{\text{Comf}} \right),
\end{aligned}
\end{equation}
where $\alpha_1,\ldots,\alpha_4$ are balancing coefficients. 

This fusion strategy jointly accounts for human preference, environmental safety, and motion smoothness, enabling the policy to generate trajectories that are both human-like and physically feasible.

\begin{table*}[htp!]
\setlength{\tabcolsep}{6pt}  % 控制列间距
\small
\center
\caption{\textbf{End-to-end planning performance on nuScenes validation set.} 
% The ego status was not utilized in the planning module.
% ``Baseline" denotes both the student and teacher models are both trained from scratch. $^\dagger$
}
\renewcommand\arraystretch{1}
\adjustbox{max width=1.0\textwidth}{
\begin{tabular}{l|c|c|ccc>{\columncolor{gray!10}}c|ccc>{\columncolor{gray!10}}c}
\toprule
\multirow{2}{*}{Method} & \multirow{2}{*}{Input} & \multirow{2}{*}{Auxiliary Supervision} & \multicolumn{4}{c|}{L2 (m) $\downarrow$} & \multicolumn{4}{c}{Collision (\%) $\downarrow$} \\
& & & {1s} & {2s} & {3s} & {Avg.} & {1s} & {2s} & {3s} & {Avg.} \\
\midrule
IL~\cite{ratliff2006maximum} & LiDAR & None & 0.44 & 1.15 & 2.47 & 1.35 & 0.08 & 0.27 & 1.95 & 0.77 \\
NMP~\cite{zeng2019end} & LiDAR & Box \& Motion & 0.53 & 1.25 & 2.67 & 1.48 & 0.04 & 0.12 & 0.87 & 0.34 \\
FF~\cite{hu2021safe} & LiDAR & Freespace & 0.55 & 1.20 & 2.54 & 1.43 & 0.06 & 0.17 & 1.07 & 0.43 \\
EO~\cite{khurana2022differentiable} & LiDAR & Freespace & 0.67 & 1.36 & 2.78 & 1.60 & 0.04 & 0.09 & 0.88 & 0.33 \\
\midrule
BevFormer \cite{li2022bevformer}+OccWorld \cite{zheng2023occworld} & Camera & 3D-Occ & 0.43 & 0.87 & 1.31 & 0.87 & - & - & - & - \\
BevFormer~\cite{li2022bevformer}+Occ-LLM~\cite{xu2025occllm} & Camera & 3D-occ & 0.26 & 0.67 & 0.98 & 0.64 & - & - & - & -\\
\midrule
ST-P3~\cite{Hu2022STP3} & Camera & Map \& Box \& Depth & 1.33 & 2.11 & 2.90 & 2.11 & 0.23 & 0.62 & 1.27 & 0.71  \\
UniAD~\cite{Hu2022uniad} & Camera & Map \& Box \& Motion \& Tracklets \& Occ & 0.48 & 0.96 & 1.65 & 1.03 & {0.05} & {0.17} & 0.71 & 0.31 \\
VAD-Base~\cite{Jiang2023vad} & Camera & Map \& Box \& Motion & {0.54} & {1.15} & {1.98} & {1.22} & 0.04 & 0.39 & 1.17 & 0.53 \\
UAD~\cite{guo2024end} & Camera & Box & 0.39 & 0.81 & 1.50 & 0.90 & 0.01 & 0.12 & 0.43 & 0.19 \\
PARA-Drive~\cite{weng2024paradrive} & Camera & Map \& Box \& Motion \& Tracklets \& Occ & 0.40 & 0.77 & 1.31 & 0.83 & 0.07 & 0.25 & 0.60 & 0.30 \\
\midrule
OccNet~\cite{tong2023scene} & Camera & 3D-Occ \& Map \& Box & 1.29 &2.13 &2.99 &{2.14} & 0.21 & 0.59 &1.37 &{0.72} \\
GenAD \cite{zheng2024genad} & Camera & Map \& Box \& Motion & 0.36 & 0.83 & 1.55 & 0.91 & 0.06 & 0.23 & 1.00 & 0.43 \\
Epona~\cite{li2024navigation} & Camera & None &  0.61 & 1.17 & 1.98 & 1.25 & 0.01 & 0.22 & 0.85 & 0.36 \\
SSR~\cite{li2024navigation} & Camera & None &  \textbf{0.24} & 0.65 & 1.36 & 0.75 & \textbf{0.00} & 0.10 & 0.36 & 0.15 \\
OccWorld \cite{zheng2023occworld} & Camera & None & 0.43 & 1.08 & 1.99 & 1.17 & 0.07 & 0.38 & 1.35 & 0.60 \\
RenderWorld \cite{yan2025renderworld} & Camera & None & 0.48 & 1.30 & 2.67 & 1.48 & 0.14 & 0.55 & 2.23 & 0.97 \\
GaussianAD \cite{zheng2024gaussianad} & Camera & 3D-Occ \& Map \& Box & 0.40 & 0.66 & 0.92 & 0.66 & 0.49 & 0.38 & 0.61 & 0.49 \\
GaussianAD \cite{zheng2024gaussianad} & Camera & 3D-Occ \& Map \& Box \& Motion & 0.40 & 0.64 & \textbf{0.88} & 0.64 & 0.09 & 0.38 & 0.81 & 0.42 \\
Drive-OccWorld \cite{Yang2024driveoccworld} & Camera & 4D-Occ & 0.32 & 0.75 & 1.49 & 0.85 & 0.05 & 0.17 & 0.64 & 0.29 \\
\midrule
Baseline & Camera & 3D-Occ & 0.29 & 0.79 & 1.56 & 0.88 & 0.70 & 0.76 & 1.71 & 1.06 \\
% Baseline & Camera & 4D-Occ & 0.27 & 0.67 & 1.23 & 0.72 & 0.68 & 0.66 & 0.80 & 0.71 \\

\rowcolor{lBlue} \textbf{WPT-Student (Ours)} & Camera & 4D-Occ  & \textbf{0.24} & 0.58  & 1.17  & 0.66  & 0.14          & 0.16 & 0.42    & {0.24}                                    \\% $^\dagger$
\rowcolor{lBlue} \textbf{WPT-Teacher (Ours)} & Camera                 & 4D-Occ                           & 0.25          & \textbf{0.58} & 1.01          & \textbf{0.61}                            & 0.16          & \textbf{0.08}          & \textbf{0.10} & \textbf{0.11}                             \\
\bottomrule
\end{tabular}}
\label{tab:nus_end2end}
\end{table*}

\subsection{World Knowledge Distillation}
\label{sec:distillation}

To enable real-time inference without relying on the world model, we transfer the teacher policy (T) reasoning ability into a lightweight student policy (S) through two distillation strategies as shown in Fig.~\ref{fig:wpt}: Policy Distillation and World Reward Distillation. Both distillation strategies are applied only during training. After training, the teacher policy and the world model are removed, leaving a compact, real-time AD policy.

\noindent\textbf{Policy Distillation.}
The teacher policy produces a set of planning queries $Q^T$ that encode future-aware reasoning guided by the world model.
The student policy generates its own set of queries $Q^S$. 
To align the student's planning intent with the teacher’s world-conditioned reasoning, we minimize the L2 distance between the two query sets:
\begin{equation}
\small
\label{eq:policy_distill}
    \mathcal{L}_{\text{policy}} = \| Q^{S} - Q^{T} \|_2.
\end{equation}
This alignment transfers the structured planning intent from the teacher into the student, allowing the student to inherit future-aware reasoning without autoregressive rollout.

\noindent\textbf{World Reward Distillation.}
The teacher generates multi-modal candidate trajectories $\mathcal{T}^T = \{\tau^T_1,\ldots,\tau^T_N\}$, which are evaluated by the reward model to obtain the optimal trajectory $\tau^*_T$ based on the final reward score (see Sec.~\ref{sec:reward_model}). 
Meanwhile, the student generates a single trajectory $\tau_S$, which is also evaluated by the reward model.
To transfer the world-model-based evaluation knowledge, we minimize the difference between the reward scores of the student’s trajectory and the teacher’s best trajectory:
% The teacher generates multi-modal trajectories $\mathcal{T}^T = \{\tau^T_1,\ldots,\tau^T_N\}$, which are evaluated by the reward model to determine the best trajectory $\tau^*_T$ under predicted future world evolution as Eq. \ref{eq:argmax_traj}.
% The student generates a single trajectory $\tau_S$, which is also scored by the reward model.
% To ensure that the student learns to produce trajectories that are preferred in the world-model-based evaluation, we minimize the reward gap:
\begin{equation}
\small
\label{eq:reward_distill}
    \mathcal{L}_{\text{reward}} = \\\| r_{\text{final}}(\tau_{S}) - r_{\text{final}}(\tau^{*}_T) \|_2.
\end{equation}
This formulation allows the student to implicitly learn the teacher’s world-informed decision criteria, effectively bridging the gap between explicit world-model reasoning and lightweight policy inference.

\section{Experiments}
\label{sec:experiment}

\subsection{Datasets and Metrics}

\noindent\textbf{NuScenes (open-loop).}
We evaluate open-loop planning on nuScenes~\cite{caesar2020nuscenes}, which contains 1{,}000 driving scenes, each 20s long, collected with a full sensor suite providing $360^\circ$ coverage. The dataset includes $\sim$1.4M images and 3D bounding boxes for 23 classes annotated at 2Hz keyframes; semantic maps are available. We follow the standard split of 700/150/150 scenes for train/val/test. Following existing practice on nuScenes~\cite{Hu2022uniad,Jiang2023vad}, we report {L2} displacement error and {collision} rate for planning quality.

\begin{table*}[htp!]
\footnotesize
\setlength{\tabcolsep}{6pt}  % 控制列间距
\center
\caption{\textbf{Open-loop and closed-loop planning performance on Bench2Drive.} Avg. L2 is averaged over the predictions in 2 seconds under 2Hz. * denotes expert feature distillation.}
% \vspace{-2mm}
\renewcommand\arraystretch{1}
\adjustbox{max width=1.0\textwidth}{
    % 在这里放表格
    % hline使用 \toprule \midrule \bottomrule代替，会让竖线出现一个小的间距更好看一些
\begin{tabular}{l|c|cccc} 
\toprule
Method           & Avg. L2 (m) $\downarrow$ & Driving Score $\uparrow$ & Success Rate (\%) $\uparrow$ & Efficiency $\uparrow$ & Comfortness $\uparrow$  \\ 
\midrule
AD-MLP~\cite{Zhai2023admlp}           & 3.64                & 18.05                   & 0.00                        & 48.45                     & 22.63                  \\
UniAD-Base~\cite{Hu2022uniad}       & 0.73                & 45.81                   & 16.36                       & 129.21                    & 43.58                  \\
UniAD-Tiny~\cite{Hu2022uniad}       & 0.80                & 40.73                   & 13.18                       & 123.92                    & 47.04                  \\
VAD-Base~\cite{Jiang2023vad}         & 0.91                & 42.35                   & 15.00                       & 157.94                    & 46.01                  \\
VAD-Tiny~\cite{Jiang2023vad}         & 1.15                & 34.28                   & 10.45                       & 70.04                     & \textbf{66.86}                  \\
SparseDrive~\cite{2025sparsedrive}      & 0.87                & 44.54                   & 16.71                       & 170.21                    & 48.63                  \\
GenAD~\cite{zheng2024genad}            & -                   & 44.81                   & 15.90                       & -                         & -                      \\
DiFSD~\cite{Su2024DiFSD}            & 0.70                & 52.02                   & 21.00                      & 178.30                    & -                      \\
DriveTransformer~\cite{2025drivetransformer} & \textbf{0.62}                & 63.46                   & 35.01                       & 100.64                    & 20.78                  \\
DiffAD~\cite{Wang2025DiffAD}           & -                   & 67.92                   & 38.64                       & -                         & -                      \\
WoTE~\cite{Li2025wote}             & -                   & 61.71                   & 31.36                       & -                         & -                      \\
DriveDPO~\cite{Shang2025DriveDPO}         & -                   & 62.02                   & 30.62                       & 166.80                    & 26.79                  \\
BridgeAD~\cite{Zhang2025bridgead}         & 0.71                & 50.06                   & 22.73                       & -                         & -                      \\
\midrule
Baseline        & 0.79                & 65.23                   & 34.10                       & 184.86                    & 23.44                  \\ 
\rowcolor{lBlue}\textbf{WPT-Student (Ours)}        & 0.75                & 72.61                   & 45.45                       & 188.52                    & 17.80                  \\ %$^\dagger$
\rowcolor{lBlue}\textbf{WPT-Teacher (Ours)}        & 0.76                & \textbf{79.23}                   & \textbf{54.54}                       & \textbf{188.63}                    & 16.39                  \\ 
\midrule
TCP-traj*~\cite{Wu2022TCP}        & 1.70                & 59.90                   & 30.00                       & 76.54                     & 18.08                  \\
ThinkTwice*~\cite{Jia2023thinktwice}      & 0.95                & 62.44                   & 31.23                       & 69.33                     & 16.22                  \\
DriveAdapter*~\cite{Jia2023DriveAdapter}    & 1.01                & 64.22                   & 33.08                       & 70.22                     & 16.01                  \\
\bottomrule
\end{tabular}
}
\label{tab:bench2drive-1}
% \vspace{-4mm}
\end{table*}

\noindent\textbf{Bench2Drive (closed-loop).}
We use Bench2Drive~\cite{jia2024bench2drive}, a CARLA-based~\cite{dosovitskiy2017carla} benchmark designed for multi-ability closed-loop E2E AD assessment, covering diverse driving scenarios. Following the official protocol, we train on 1000 clips (950 for training and 50 for open-loop validation) and compare closed-loop results on the predefined 220 routes.

\subsection{Implementation Details}

\noindent\textbf{Training on NuScenes Setting.}
We adopt the pre-trained Drive-OccWorld~\cite{Yang2024driveoccworld} model as our world model and freeze its weights during training. 
WPT is trained for 12 epochs with a batch size of 8, using AdamW and a cosine annealing learning rate schedule.

\noindent\textbf{Training on Bench2Drive Setting.}
% For experiments on the Bench2Drive benchmark, 
We adopt an instance-based world model that forecasts future agent states and lane topology instead of occupancy (detailed in our appendix).
WPT is trained for 6 epochs with a batch size of 16, also using AdamW and a cosine annealing learning rate schedule.

\subsection{Main Results}

\noindent\textbf{Open-Loop Results on NuScenes.}
We compare WPT with recent end-to-end planners on nuScenes, shown in~\cref{tab:nus_end2end}. As a reference, our Baseline uses the same student structure without the proposed reward or distillation. Against this baseline, WPT-Teacher improves Avg. L2 to 0.61m and collision to 0.11\%, while WPT-Student retains most of the gains without test-time world-model overhead.% ({–30.7\%}) ({–89.6\%}) 
Compared with strong world-model methods, WPT-Teacher also achieves the best Avg. L2 and the lowest collision (\eg, vs. Drive-OccWorld~\cite{Yang2024driveoccworld}: 0.85m, 0.29\%).
By horizon, WPT-Teacher leads at 2s (0.58m) and markedly reduces 3s collisions to 0.10\%, indicating stronger long-horizon foresight, benefiting from our world-aware training.
Importantly, our lightweight WPT-Student preserves these performance improvements while achieving world-model-free inference.
These results validate that training-time interaction and world-aware distillation transfer predictive awareness into a compact policy that effectively improves the student performance. 

%% As a reference point, our \emph{Baseline} (same student architecture, no rewards/distillation) reaches 
\noindent\textbf{Closed-Loop Results on Bench2Drive.}
We evaluate WPT on Bench2Drive~\cite{jia2024bench2drive} under both open-loop (Avg.\ L2) and closed-loop metrics, as summarized in~\cref{tab:bench2drive-1}. Additional multi-modality analyses across diverse scenarios are provided in our appendix. 
WPT-Teacher achieves the best {Driving Score} (79.23), {Success Rate} (54.54\%), and {Efficiency} (188.63), surpassing recent E2E planners such as DriveTransformer~\cite{2025drivetransformer} (63.46 DS / 35.01\% SR / 100.64 Eff.) and DriveDPO~\cite{Shang2025DriveDPO} (62.02 DS / 30.62\% SR / 166.80 Eff.). 
Relative to the {Baseline} (same student architecture, no rewards/distillation), WPT-Teacher boosts DS by +14.00 and SR by +20.44 points and the distilled {WPT-Student} retains most of the gains.
Despite the lower open-loop Avg. L2 (0.62m) of DriveTransformer, both WPT variants yield stronger closed-loop results, consistent with Bench2Drive’s emphasis on closed-loop ability.
Compared with methods that rely on expert feature distillation, our models deliver higher DS and SR without requiring expert knowledge.
We note a known efficiency–comfort trade-off; improving smoothness without compromising efficiency and success is left for future reward-shaping refinements.
Overall, the gains indicate that our interaction-based rewards and world-aware distillation yield a policy that is effective in closed-loop rollouts.
% and aligned with closed-loop multi-ability evaluation protocol. 

\subsection{Ablation Study}
To validate each design in our method, we conduct comprehensive studies on nuScenes dataset using average L2 error and collision rate as planning metrics. Baseline-T denotes the teacher planner WPT trained without our reward model.

\noindent\textbf{Effect of Reward Model.}
We evaluate when interaction-based rewards are applied, during training and at inference, as shown in~\cref{tab:reward}. Starting from Baseline-T (0.72m / 0.71\%), adding the imitation reward in training reduces collisions to 0.23\% (0.70m L2). Applying the same reward at inference leads to better results (0.69m / 0.22\%). Adding simulation rewards gives the largest gains: training-only reaches 0.62m / 0.14\%, and enabling inference-time scoring attains the best 0.61m / 0.11\%. The results show the effectiveness of our reward model. Meanwhile, we find that most benefits come from training-time supervision; optional inference-time scoring adds a small margin at the cost of runtime coupling (see our appendix).
% Baseline-T: (teacher) driveoccworld freeze, only train planning
% +imi: 
% +imi+sim: 
\begin{table}[h]
    \footnotesize
    \setlength{\tabcolsep}{0.01\linewidth}
    \centering
    \caption{\textbf{Ablation study of reward model.} We compare different reward equipment at different usage stages (training stage or also at inference). ``Im. Rwd." is an imitation reward, while ``Sim.Rwd." means simulation reward.}
    % \vspace{-3mm}
    \adjustbox{max width=1.0\linewidth}{
    \begin{tabular}{l|cc|cc} 
    \toprule
    \multirow{2}{*}{\textbf{Method}}      & \multicolumn{2}{c|}{\textbf{Stage}}                             & \multicolumn{2}{c}{\textbf{Planning}}  \\
                                 & Train                         & Infer.                     & Avg. L2(m)$\,\downarrow$   & Avg. Col.(\%)$\downarrow$          \\ 
    \midrule
    Baseline-T                     & -                             & -                             & 0.72 & 0.71               \\ 
    \midrule
    \multirow{2}{*}{+ Im. Rwd.} & $\checkmark$ & -                             & 0.78  & 0.23               \\
                                 & $\checkmark$ & $\checkmark$ & 0.69 & 0.22               \\ 
    \midrule
    \multirow{2}{*}{+ Im.\&Sim. Rwd.} & $\checkmark$ & -                             & 0.62 & 0.14               \\
                                 & $\checkmark$ & $\checkmark$ & \textbf{0.61} & \textbf{0.11}               \\
    \bottomrule
    \end{tabular}
}
    \label{tab:reward}
% \vspace{-6mm}
\end{table}
%
% 6个 reward head
% im
% 5 个 sim: NC, DAC, EP, TTC, Comf.
\begin{table}[h]
% 控制列间距
\small
\setlength{\tabcolsep}{0.01\linewidth}
\centering
\caption{\textbf{Ablation study of different rewards.} Simulation reward consists of five signals: NC (No Collision), EP (Ego Progress), DAC (Drivable Area Compliance), TTC (Time-to-Collision), and Comf. (Comfort).}% NC, DAC, EP, TTC, Comf. %are refer to the navisim dataset.}
% \vspace{-2mm}
\adjustbox{max width=0.86\linewidth}{
    % 在这里放表格
    % hline使用 \toprule \midrule \bottomrule代替，会让竖线出现一个小的间距更好看一些
    \begin{tabular}{c|ccccc|cc} 
    \toprule
    % \multicolumn{6}{c|}{Reward Type}                                                        & \multicolumn{2}{c}{Planning}  \\
    \multirow{2}{*}{\textbf{Im. Rwd.}} & \multicolumn{5}{c|}{\textbf{Sim. Rwd.}} & \multicolumn{2}{c}{\textbf{Planning}} \\
    % \cmidrule(lr){2-6} \cmidrule(lr){7-8} 
     & NC & DAC & EP & TTC & Comf. & L2 (m)$\downarrow$ & Col. (\%)$\downarrow$ \\
    % Im.    & NC           & DAC          & EP           & TTC          & Comf.        & Avg. L2 (m)$\downarrow$ & Avg. Col. (\%)$\downarrow$  \\ 
    \midrule
    -            & -            & -            & -            & -            & -            & 0.72        & 0.71            \\ 
    
    $\checkmark$ & -            & -            & -            & -            & -            & 0.76        & 0.23            \\
    $\checkmark$ & $\checkmark$ & $\checkmark$ & $\checkmark$ & $\checkmark$ & $\checkmark$ & \textbf{0.61}        & \textbf{0.11}            \\
    % \midrule
    $\checkmark$ & -            & $\checkmark$ & $\checkmark$ & $\checkmark$ & $\checkmark$ & 0.62        & 0.22            \\
    $\checkmark$ & $\checkmark$ & -            & $\checkmark$ & $\checkmark$ & $\checkmark$ & 0.62        & 0.23            \\
    $\checkmark$ & $\checkmark$ & $\checkmark$ & -            & $\checkmark$ & $\checkmark$ & 0.62        & 0.23            \\
    $\checkmark$ & $\checkmark$ & $\checkmark$ & $\checkmark$ & -            & $\checkmark$ & 0.69        & 0.25            \\
    $\checkmark$ & $\checkmark$ & $\checkmark$ & $\checkmark$ & $\checkmark$ & -            & 0.63        & 0.23            \\
    \bottomrule
    \end{tabular}
}
\label{tab:reward_type}
% \vspace{-2mm}
\end{table}

\noindent\textbf{Effect of Different Rewards.}
We analyze the composition of our reward signals, as shown in~\cref{tab:reward_type}. With imitation reward only, performance is 0.76m / 0.23\%. Further aggregating all simulation rewards (NC, DAC, EP, TTC, Comf.) achieves the best 0.61m / 0.11\%, indicating complementary effects. Ablating each simulation reward degrades performance; removing the TTC reward signal causes the largest collision increase (to 0.25\%), highlighting its importance for safety, while other simulation reward signals contribute smaller but steady gains.
%to both L2 and collision.

\noindent\textbf{Effect of Interaction Sources.}
We compare using ground-truth occupancy (GT-Occ) versus world-model rollouts (WM-Occ) to drive reward signals, shown in~\cref{tab:sim_type}. With imitation reward only, GT-Occ offers stronger signals (0.64m / 0.16\% vs.\ 0.69m / 0.22\%). With equipping simulation rewards, WM-Occ achieves the best overall (0.61m / 0.11\%) compared to GT-Occ (0.65m / 0.11\%), suggesting that training on world-model rollouts with WM-Occ aligns the policy with the predictive structure compared to the deterministic GT-Occ.

\noindent\textbf{Effect of Distillation Strategies.}
We study the lightweight student with different distillation strategies, shown in~\cref{tab:distill_type}. Plain training yields 0.88m / 1.06\%. Query-level distillation alone provides a sizable step (0.69m / 0.86\%). Adding imitation-reward distillation sharply reduces collisions to 0.25\% (0.68m). Combining imitation and simulation reward distillation attains the best student performance, 0.66m / 0.24\%, demonstrating that world-aware signals effectively transfer predictive awareness without test-time world-model calls.

\begin{table}[t]
% 控制列间距
\setlength{\tabcolsep}{0.01\linewidth}
\small
\centering
\caption{\textbf{Interaction occupancy source ablation.} GT-Occ denotes using ground truth occupancy for interaction, while WM-Occ denotes using the occupancy generated by WM.}
% \vspace{-2mm}
\adjustbox{max width=0.86\linewidth}{
    % 在这里放表格
    % hline使用 \toprule \midrule \bottomrule代替，会让竖线出现一个小的间距更好看一些
    \begin{tabular}{l|cc|cc} 
    \toprule
    \multirow{2}{*}{\textbf{Method}}      & \multicolumn{2}{c|}{\textbf{Interaction Source}} & \multicolumn{2}{c}{\textbf{Planning}}  \\
                                 & GT-Occ       & WM-Occ           & L2 (m)$\downarrow$ & Col. (\%)$\downarrow$    \\ 
    \midrule
    Baseline-T                     & -            & $\checkmark$     & 0.72       & 0.71             \\ 
    \midrule
    \multirow{2}{*}{+ Im. Rwd.}  & $\checkmark$ & -                & 0.64       & 0.16             \\
                                 & -            & $\checkmark$     & 0.69       & 0.22             \\ 
    \midrule
    \multirow{2}{*}{+ Im.\&Sim. Rwd.} 
    & $\checkmark$ & -                & 0.65       & 0.11             \\
                                 & -            & $\checkmark$     & \textbf{0.61}       & \textbf{0.11}             \\
                                 % & $\checkmark$ & -                & 0.65       & 0.11             \\
    \bottomrule
    \end{tabular}
}
\label{tab:sim_type}
% \vspace{-3mm}
\end{table}

\begin{table}[t]
% 控制列间距
\setlength{\tabcolsep}{0.01\linewidth}
\centering
\caption{\textbf{Ablation study of distillation.} ``Query" denotes the distillation of the plan query. 
% ``Im.Rwd.": imitation reward distillation; ``Sim.Rwd": simulation reward distillation.
}
% \vspace{-2mm}
\adjustbox{max width=0.86\linewidth}{
    % 在这里放表格
    % hline使用 \toprule \midrule \bottomrule代替，会让竖线出现一个小的间距更好看一些
    \begin{tabular}{ccc|cc} 
    \toprule
    \multicolumn{3}{c|}{\textbf{Distillation Type}}    & \multicolumn{2}{c}{\textbf{Planning}}  \\
    Query        & Im. Rwd.  & Sim. Rwd. & Avg. L2 (m) $\downarrow$   & Avg. Col. (\%) $\downarrow$                   \\ 
    \midrule
    -            & -            & -           & 0.88 & 1.06                   \\
    $\checkmark$ & -            & -           & 0.69 & 0.86                   \\
    $\checkmark$ & $\checkmark$ & -           & 0.68 & 0.25                   \\
    $\checkmark$ & $\checkmark$  & $\checkmark$ & \textbf{0.66} & \textbf{0.24}                   \\
    \bottomrule
    \end{tabular}
}
\label{tab:distill_type}
% \vspace{-4mm}
\end{table}
\noindent \textbf{Computation and Latency.} 
We report full training compute (GPU hours) and planning inference latency in the appendix. Briefly, the distilled {WPT-Student} matches Baseline latency (64\,ms) and is $4.9\times$ faster than {WPT-Teacher} (312\,ms), while improving planning from 0.88m / 1.06\% to 0.66m / 0.24\%.

Overall, these ablations show that: 1) interaction-based rewards are the primary driver of safety gains; 2) simulator reward signals provide complementary improvements; 3) supervision from world-model rollouts is preferable to deterministic GT; 4) world-aware distillation consolidates these effects into a compact, deployable planner.

\section{Conclusion}
\label{sec:conclusion}

% In this paper, we present \textbf{WPT}, a World-to-Policy Transfer training paradigm that bridges world model and end-to-end policy learning. 
% Unlike previous approaches that depend on runtime interaction with the world model or offline reward computation, WPT performs online knowledge transfer during training and achieves real-time inference without relying on the world model. 
% By introducing a trainable reward model and two distillation schemes, namely policy distillation and world reward distillation, our framework effectively infuses world-model knowledge into a lightweight policy network. 
% Extensive experiments on both open-loop and closed-loop benchmarks demonstrate that WPT achieves state-of-the-art performance in planning accuracy, driving safety, and efficiency. 
% We believe this work provides a new perspective on integrating world modeling and E2E AD, paving the way toward more scalable, interpretable, and real-time driving systems.

In this paper, we present \textbf{WPT}, a World-to-Policy Transfer training paradigm for end-to-end AD. Through a trainable reward model and dual distillation schemes, WPT distills world-model knowledge into a lightweight policy during training. This eliminates runtime world-model dependencies, ensuring real-time inference. Extensive experiments on both open-loop and closed-loop benchmarks demonstrate that WPT achieves state-of-the-art performance in planning accuracy, driving safety, and efficiency.

\section*{Acknowledgements}
This work was supported in part by the National Natural Science Foundation of China under Contract 62471450, and the Natural Science Foundation of Anhui Province under Grant 2208085J17.

% \clearpage

{
    \small
    \bibliographystyle{ieeenat_fullname}
    \bibliography{main}
}

\setcounter{page}{1}
\maketitlesupplementary

% Image example
% \begin{figure}[t]
%     \centering
%     \includegraphics[width=0.96\linewidth]{images/introduction_compare.pdf}
%     \vspace{-3mm}
%     \caption{\textbf{Your Figure Caption.}.}
%     \vspace{-4mm}
%     \label{fig:intro_compare}
% \end{figure}

\section{More Experiment Details}
% \tableofcontents
\begin{figure*}[h!]
    \centering
    \includegraphics[width=0.98\linewidth]{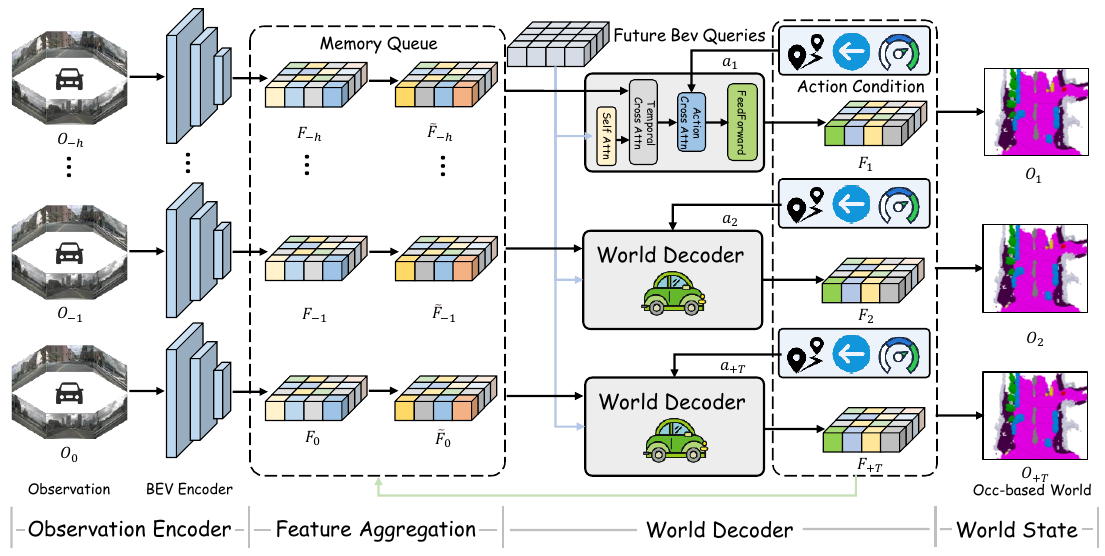}
    % \vspace{-6mm}
    \caption{\textbf{Detailed structure of the occupancy-based world model,} which predicts the future world states through an autoregressive manner. The model utilizes an observation encoder to process multi-view images, a feature aggregation module to capture temporal consistency, and a world decoder to predict the future BEV embedding based on historical and current world features. This approach allows the model to predict future occupancy states.}
    % \vspace{-4mm}
    \label{fig:occ_wm}
\end{figure*}

\subsection{World Model Structure}
% 在这一部分我们详细介绍了我们使用的两个世界模型，如图4和5所示。我们在NuScenes数据集上使用基于Occupancy的世界模型~\cite{driveoccworld}，在Bench2Drive上使用基于Instance的世界模型~\cite{VAD}。

% Occupancy-based 世界模型
% 如图4所示，世界解码器基于Memory Queue中存储的历史BEV特征和预期动作条件，采用自回归方式预测下一时间戳的BEV嵌入。 具体而言，可学习的未来BEV query首先通过自注意力机制建立上下文关联，随后采用时间交叉注意力层从多帧历史嵌入中提取对应特征。  接着，条件交叉注意力层在BEV查询与条件嵌入之间进行交叉注意力交互，将动作条件注入预测过程。最终，前馈网络输出生成的BEV特征，可用于未来占用率。

% Instance-based 世界模型
% 如图5所示，我们的Instance-based世界模型是一个基于VAD的魔改世界模型。多视图图像通过基于ResNet的编码器进行特征提取。随后，基于BEVFormer的解码器将这些特征转换为BEV空间的特征。为了表示世界的未来演变，我们预测未来的静态道路拓扑结构（例如车道标线、车道中心线、人行道及其他道路结构）和Agent motion，通过解码对应的Query得到。

\begin{figure*}[h!]
    \centering
    % \vspace{-8mm}
    \includegraphics[width=0.98\linewidth]{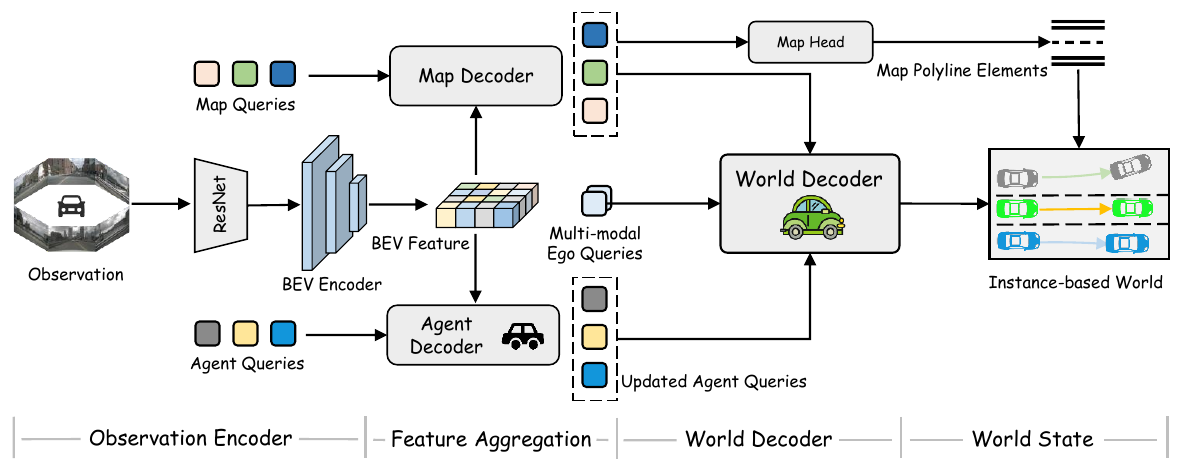}
    % \vspace{-8mm}
    \caption{\textbf{Detailed structure of the instance-based world model,} which predicts the future map elements and agent motion. The model uses a ResNet encoder to process multi-view images, a BEV encoder to transform the features into BEV space, and two decoders: the map decoder for static road elements and the world decoder for dynamic agents.}
    % \vspace{-4mm}
    \label{fig:instance_wm}
\end{figure*}

In this section, we provide a detailed description of the world model used in our WPT. As shown in \cref{fig:occ_wm,fig:instance_wm}, we utilize an occupancy-based world model~\cite{Yang2024driveoccworld} on the nuScenes dataset and an instance-based world model, which is similar to VAD~\cite{Jiang2023vad} on the Bench2Drive dataset.

% \noindent \textbf{Long-Tailed Small Objects.}

\begin{figure*}[h!]
    \includegraphics[width=0.98\linewidth]{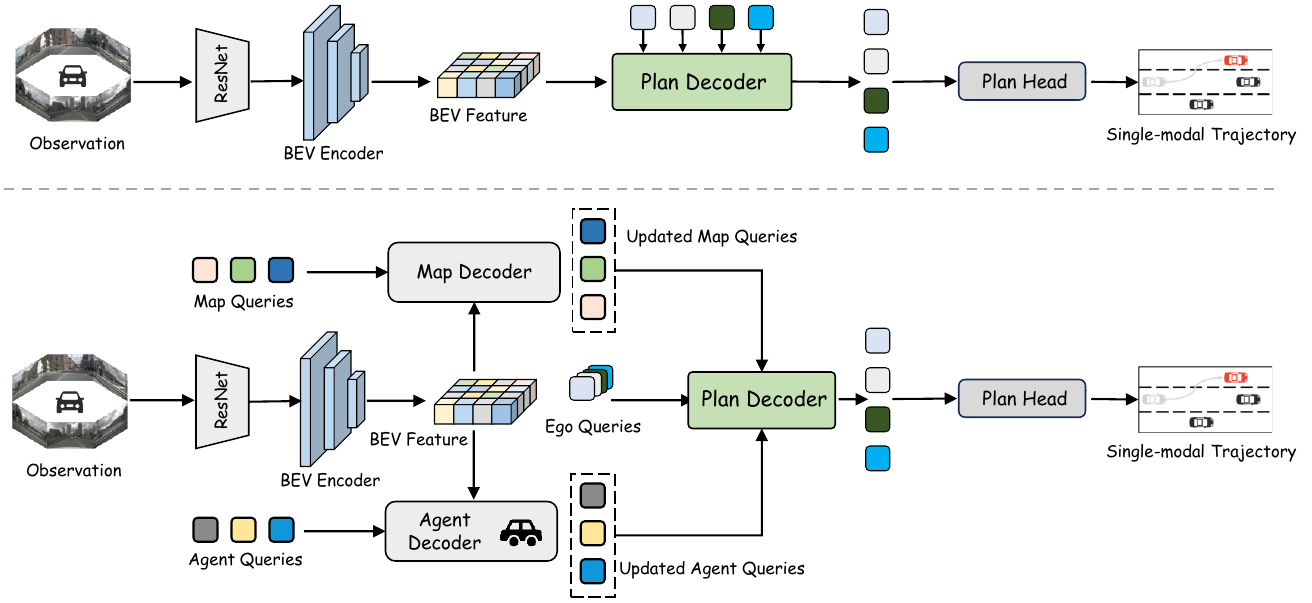}
    \caption{\textbf{Illustration of our Occ-based and instance-based baseline models.} The top part shows the occupancy-based baseline model, while the bottom part illustrates the instance-based baseline model. Both approaches utilize a BEV decoder, but differ in how planning queries interact with features.}
    \label{fig:baseline}
\end{figure*}

\noindent \textbf{Occupancy-based World Model.} The structure of the occupancy-based world model is illustrated in \cref{fig:occ_wm}. The world decoder predicts the future BEV embedding based on the historical BEV features stored in the memory queue and the expected action conditions using an autoregressive approach. Specifically, the future BEV queries first establish contextual associations through a self-attention mechanism. Then, a temporal-cross attention layer extracts the corresponding features from multiple historical embeddings. Following this, an action cross-attention layer enables interaction between BEV queries and action conditions, injecting the action context into the prediction process. Finally, the feed-forward network generates the predicted BEV features, which are used for future occupancy predictions.

\noindent \textbf{Instance-based World Model}
As illustrated in~\cref{fig:instance_wm}, our instance-based world model is based on a modified version of VAD~\cite{Jiang2023vad}. In this model, multi-view images are first processed by a ResNet-based encoder to extract features. These features are then decoded into BEV space using a BEVFormer-based decoder.
To predict the future world evolution, we model both the static road topology (\eg., lane markings, lane centerlines, sidewalks, and other road structures) and dynamic agent motion. The future state is predicted by decoding the corresponding queries for both static elements and dynamic agents.

% \subsection{WPT-Baseline}
% 对于Occupancy-based world model，我们的Baseline的结构如图所示，通过img_backbone,img_neck提取环视图特征，再通过一个基于bevformer的bev decoder得到bev特征。然后，过初始化的Plan Query与提取的bev特征进行交互，得到未来的Refined Plan Query，再通过一个plan head直接得到对应未来的多帧轨迹。
% 对于instance-based world model，我们的Baseline结构是基于图5的instance-based世界模型进行修改的，Mulit-modal ego queries与map queries和agent queries进行cross attention得到refined ego queries，然后通过一个plan head直接解码出未来的多帧轨迹。

\subsection{WPT-baseline Models}
% 对于occupancy-based world mode1，我们的Baseline的结构如图所示，通过img_backbone,img_neck提取环视图特征，再通过一个基于bevformer的bev decoder得到bev特征。然后，过初始化的P1an Query与提取的bev特征进行交互，得到未来的Refined plan Query，再通过一个plan head直接得到对应未来的多帧轨迹。
% 对于instance-based world mode1，我们的Baseline结构是基于图5的instance-based世界模型进行修改的，Mulit-modal ego queries与map queries和agent queries进行cross attention得到refined ego queries，然后通过一个p1an head直接解码出未来的多帧轨迹。

\noindent\textbf{Occupancy-based Baseline.} The architecture of the occupancy-based baseline planner is shown at the top of~\cref{fig:baseline}. The model extracts BEV features through the image backbone and BEV Encoder~\cite{li2022bevformer} from multi-view input images, which are then processed by a BEV decoder~\cite{li2022bevformer}. The initialized planning query interacts with these extracted BEV features, producing a refined plan query that predicts future trajectories. 

\noindent\textbf{Instance-based Baseline.} The architecture of our instance-based baseline planner is shown in the bottom part of~\cref{fig:baseline}. Here, ego queries interact with map and agent queries via cross-attention, refining the ego query. These refined queries are then decoded by the plan head to generate future trajectory predictions.

\subsection{Simulation Reward Design}
\label{sec:simulation_reward_label}
% 在这章节，我们提供了详细的基于世界模型的模拟奖励(NC (No Collision), EP (Ego Progress), DAC (Drivable Area Compliance), TTC (Time-to-Collision) and Comfort)的指标计算。
% Occupancy-Based
% NC. 
% DAC.
% TTC
% Comfort
In this section, we provide a detailed design of the Simulation Reward mechanism, which includes five distinct reward components: no collision (NC), drivable area compliance (DAC), ego progress (EP), time-to-collision (TTC), and comfort (Comf). These rewards assess the safety, efficiency, and comfort of the generated trajectories from an environment-centered perspective. Each reward is calculated based on the interaction of the candidate trajectory with the predicted world features, such as obstacles, drivable areas, and the ego's progress.

Below is the detailed description of how each reward is calculated and incorporated into the model.

\noindent\textbf{1. NC.}  
The NC reward evaluates whether the candidate trajectory intersects with obstacles in the environment. The trajectory points are compared against the predicted occupancy grid (instance occupancy) to identify collisions. If no collision occurs along the trajectory, the reward is maximized (i.e., a score of 1), and if a collision is detected, the reward is minimized (i.e., a score of 0). The final NC reward is computed as:
\begin{equation}
\small
\label{eq:nc_reward}
    S_{\text{NC}}(\tau) = \begin{cases} 
1 & \text{if no collision occurs,} \\
0 & \text{if collision occurs.}
\end{cases}
\end{equation}

\noindent\textbf{2. DAC.}  
The DAC reward ensures that the trajectory stays within the predicted drivable areas. The penalty is applied whenever any point of the trajectory moves outside the permissible drivable zone, signifying a violation of traffic rules. The reward is calculated as follows: the trajectory receives a reward of 1 if all points lie within the drivable area, and 0 if any point of the trajectory lies outside the drivable area. The final DAC reward is given by:
\begin{equation}
\small
\label{eq:dac_reward}
S_{\text{DAC}}(\tau) = \begin{cases}
1 & \text{if $\tau$ is within the drivable area,} \\
0 & \text{if $\tau$ is outside the drivable area.}
\end{cases}
\end{equation}

\noindent\textbf{3. EP.}  
The EP reward evaluates the ego's progress along the trajectory by measuring its forward movement in terms of longitudinal displacement. Positive progress is encouraged, while negative progress (backward movement) is penalized. To ensure that the trajectory remains safe, the progress reward is only activated if the NC and DAC conditions are satisfied. If either of these conditions is violated, the reward is set to zero. The reward is computed as follows:
\begin{equation}
\small
\label{eq:ep_reward}
S^{'}_{\text{EP}}(\tau) = \begin{cases}
\frac{C_{\text{EP}}(\tau)}{C_{\text{EP}}^{\text{max}}} & \text{if } C_{\text{EP}}^{\text{max}} > 5.0 \text{ and } C_{\text{EP}}(\tau) \geq 0, \\
1 & \text{if } C_{\text{EP}}^{\text{max}} \leq 5.0 \text{ and } C_{\text{EP}}(\tau) \geq 0, \\
0 & \text{otherwise},
\end{cases}
\end{equation}
\begin{equation}
\small
\label{eq:eq_reward_2}
S_{\text{EP}}\left( \tau \right) =S^{'}_{\text{EP}}\left( \tau \right) S_{\text{NC}}\left( \tau \right) S_{\text{DAC}}\left( \tau \right),
\end{equation}
where $C_{\text{EP}}(\tau)$ represents the longitudinal displacement along the trajectory $\tau$. $C_{\text{EP}}^{\text{max}}$ is the maximum longitudinal displacement of the candidate trajectory within the batch.

\noindent\textbf{4. TTC.}  
The TTC reward evaluates the ego vehicle's distance to potential obstacles in its future trajectory. To ensure safety, the trajectory is extended forward by a fixed distance \( d_{\text{fix}} = 10 \)m, and obstacles within the drivable region are checked. If no collision risk is detected within this extended range, the TTC reward is assigned a value of 1, indicating a safe trajectory. If a collision risk is detected, the reward is set to 0, indicating an unsafe trajectory. The final TTC reward is given by:
\begin{equation}
\small
\label{eq:ttc_reward_1}
S^{'}_{\text{TTC}}(\tau) = \begin{cases} 
1 & \text{if no collision risk is detected}, \\
0 & \text{if collision risk is detected},
\end{cases}
\end{equation}
\begin{equation}
\small
\label{eq:ttc_reward_2}
S_{\text{TTC}}\left( \tau \right) =S^{'}_{\text{TTC}}\left( \tau \right) S_{\text{DAC}}\left( \tau \right).
\end{equation}

\begin{table*}[t]
\setlength{\tabcolsep}{6pt}  % 控制列间距
\center
\caption{\textbf{Multi-ability performance of E2E methods.} * denotes expert feature distillation.}
\renewcommand\arraystretch{1}
\adjustbox{max width=1.0\textwidth}{
% 在这里放表格
% hline使用 \toprule \midrule \bottomrule代替，会让竖线出现一个小的间距更好看一些
\begin{tabular}{l|ccccc|>{\columncolor{gray!10}}c} 
\toprule
\multirow{2}{*}{Method} & \multicolumn{6}{c}{Ability (\%) $\uparrow$}                                                             \\
                        & Merging        & Overtaking     & Emergency Brake & Give Way       & Traffic Sign   & Mean            \\ 
\midrule
AD-MLP~\cite{Zhai2023admlp}                  & 0.00           & 0.00           & 0.00            & 0.00           & 4.35           & 0.87            \\
UniAD-Tiny~\cite{Hu2022uniad}              & 8.89           & 9.33           & 20.00           & 20.00          & 15.43          & 14.73           \\
UniAD-Base~\cite{Hu2022uniad}              & 14.10          & 17.78          & 21.67           & 10.00          & 14.21          & 15.55           \\
VAD~\cite{Jiang2023vad}                     & 8.11           & 24.44          & 18.64           & 20.00          & 19.15          & 18.07           \\
DriveTransformer~\cite{2025drivetransformer}        & 17.57          & 35.00          & 48.36           & 40.00          & 52.10          & 38.60           \\
DiffAD~\cite{Wang2025DiffAD}                  & 30.00          & 35.55          & 46.66           & 40.00          & 46.32          & 38.79           \\
\midrule
Baseline     & 26.26 & 40.00 & 40.00  & 50.00 & 40.53          & 39.36  \\ 
\rowcolor{lBlue}\textbf{WPT-Student (Ours)}               & 30.00 & 55.56 & 63.33  & 50.00 & 47.37          & 49.25  \\ 
\rowcolor{lBlue}\textbf{WPT-Teacher (Ours)}               & \textbf{47.50} & \textbf{73.33} & \textbf{65.00}  & \textbf{50.00} & \textbf{53.16}          & \textbf{57.80}  \\ 
\midrule
TCP*~\cite{Wu2022TCP}                   & 16.18          & 20.00          & 20.00           & 10.00          & 6.99           & 14.63           \\
TCP-Ctrl*~\cite{Wu2022TCP}               & 10.29          & 4.44           & 10.00           & 10.00          & 6.45           & 8.23            \\
TCP-Traj*~\cite{Wu2022TCP}               & 8.89           & 24.29          & 51.67           & 40.00          & 46.8           & 34.22           \\
ThinkTwice*~\cite{Jia2023thinktwice}             & 27.38          & 18.42          & 35.82           & 50.00          & 54.23          & 37.17           \\
DriveAdapter*~\cite{Jia2023DriveAdapter}           & 28.82          & 26.38          & 48.76           & 50.00          & \textbf{56.43} & 42.08           \\
\bottomrule
\end{tabular}
}
% \vspace{-4mm}
\label{tab:bench2drive-ability}
\end{table*}

\noindent\textbf{5. Comf.}  
The Comf.~reward evaluates the smoothness and comfort of the trajectory by penalizing abrupt longitudinal/lateral accelerations and jerk.  
Given a candidate trajectory $\tau$, we compute the longitudinal acceleration $a_{\text{lon}}$, lateral acceleration $a_{\text{lat}}$, jerk magnitude $\|\mathbf{j}\|$, and longitudinal jerk $j_{\text{lon}}$ based on the ego-vehicle's kinematic profile.  
The comfort reward is assigned only when all motion quantities fall within acceptable thresholds:
\begin{equation}
\small
\label{eq:comf_reward_1}
\begin{aligned}
S_{\text{Comf.}}(\tau) =\;& 
\mathbf{1}\left[a_{\text{lon}} \in \left[a_{\text{min}}, a_{\text{max}}\right]\right] \\
&\cdot \mathbf{1}\left[|a_{\text{lat}}| \le a_{\text{lat}}^{\text{max}}\right] \\
&\cdot \mathbf{1}\left[\|\mathbf{j}\| \le j^{\text{max}}\right] \\
&\cdot \mathbf{1}\left[|j_{\text{lon}}| \le j_{\text{lon}}^{\text{max}}\right],
\end{aligned}
\end{equation}
% where $\mathbf{1}\left[\cdot\right]$ is an indicator function that outputs $1$ if the condition is satisfied and $0$ otherwise.  
% The thresholds are defined according to NAVISIM comfort standards:
% \begin{itemize}
%     \item $a_{\text{min}} = -4.05\,\text{m/s}^2$: minimum admissible longitudinal deceleration, $a_{\text{max}} = 2.40\,\text{m/s}^2$: maximum admissible longitudinal acceleration,
%     \item $a_{\text{lat}}^{\text{max}} = 4.89\,\text{m/s}^2$: maximum admissible lateral acceleration,
%     \item $j^{\text{max}} = 8.37\,\text{m/s}^3$: maximum admissible jerk magnitude,
%     \item $j_{\text{lon}}^{\text{max}} = 4.13\,\text{m/s}^3$: maximum admissible longitudinal jerk.
% \end{itemize}
% Here,  
% - $a_{\text{lon}}$ measures acceleration along the vehicle's heading direction,  
% - $a_{\text{lat}}$ measures lateral acceleration due to steering,  
% - $\mathbf{j}$ denotes the jerk vector (time derivative of acceleration),  
% - $j_{\text{lon}}$ is the longitudinal component of jerk.  
% A trajectory is considered comfortable only when \textit{all} of these constraints are satisfied.
where $\mathbf{1}[\cdot]$ is an indicator function. The thresholds are based on NAVISIM comfort standards:
\begin{itemize}
    \item \(a_{\text{min}} = -4.05\,\text{m/s}^2\): minimum longitudinal deceleration, \(a_{\text{max}} = 2.40\,\text{m/s}^2\): maximum longitudinal acceleration,
    \item \(a_{\text{lat}}^{\text{max}} = 4.89\,\text{m/s}^2\): maximum lateral acceleration,
    \item \(j^{\text{max}} = 8.37\,\text{m/s}^3\): maximum jerk magnitude,
    \item \(j_{\text{lon}}^{\text{max}} = 4.13\,\text{m/s}^3\): maximum longitudinal jerk.
\end{itemize}
Here, \(a_{\text{lon}}\) and \(a_{\text{lat}}\) represent longitudinal and lateral accelerations, respectively, while \(\mathbf{j}\) is the jerk vector, and \(j_{\text{lon}}\) is its longitudinal component. A trajectory is considered comfortable when all constraints are satisfied.

\section{More Closed-loop Experiments}
% 表7进一步对比了不同方法在特定驾驶场景中的成功率。只有当自动驾驶车辆在不发生碰撞或违规的情况下抵达指定目的地时，该场景才被视为成功。我们的模型（WPT-Student和WPT-Teacher）在多项关键驾驶技能中始终优于其他模型，Our model consistently outperforms competitors across several critical driving skills, achieving notably high success rates in Merging (47.50%), Overtaking (73.33%), Emergency Brake (65.00%), and Give Way (50.00%), Traffic Sign (53.16%). Obtains the highest overall ability score of 57.80%,位居榜首。这些数据充分证明，我们的方法在处理复杂且高度交互的驾驶场景时，展现出强大的泛化能力。

\subsection{Multi-Ability Results}
~\cref{tab:bench2drive-ability} further compares the success rates of different methods across five challenging driving scenarios. 
A scenario is considered successful only when the ego vehicle reaches the target destination without any collisions or infractions.  

Both WPT-Teacher and WPT-Student achieve strong improvements over the Baseline across all abilities. 
In particular, WPT-Teacher obtains the highest success rates in \textit{Merging} (47.50\%), \textit{Overtaking} (73.33\%), 
\textit{Emergency Brake} (65.00\%), and \textit{Traffic Sign} (53.16\%), leading to the best overall ability score of \textbf{57.80\%}.  
Meanwhile, the lightweight WPT-Student also surpasses the Baseline notably (49.25\% vs.\ 39.36\%), demonstrating that 
our distillation strategy preserves strong multi-ability performance.  
These results highlight the effectiveness of WPT in handling complex, interactive driving scenarios with superior generalization capability.

% WPT加速训练
% baseline+（student） (driveoccworld:extact his feat, imgbackbone,imgneck,pts_bbox_head (exact bev feat) + 3.1.2 single-modal(6个plan query，6帧单模轨迹)) all train train from scratch.
% WPT+ WPT蒸馏baseline+
% 问下旭哥
% A800 8卡，bsz=8
% baseline换成driveoccworld
\begin{table}[ht]
% 控制列间距
\setlength{\tabcolsep}{0.01\linewidth}
\centering
\caption{\textbf{Comparison of training process.} }
% \vspace{-2mm}
\adjustbox{max width=1.0\linewidth}{
    % 在这里放表格
    % hline使用 \toprule \midrule \bottomrule代替，会让竖线出现一个小的间距更好看一些
    \begin{tabular}{l|cc|cc}
    \hline
    \multirow{2}{*}[-1pt]{\textbf{Method}} & \multicolumn{2}{c|}{\textbf{Training Settings}} & \multicolumn{2}{c}{\textbf{Planning}}  \\
    & Epoch & GPU Hours (h)                & Avg. L2 (m) $\downarrow$ & Avg. Col. (\%) $\downarrow$ \\ 
    \toprule
    Baseline-T                & 24    & 464                          & 0.72        & 0.71            \\
    WPT-Teacher                     & 12    & 248                          & 0.61        & 0.11            \\ 
    \midrule
    Baseline               & 24    & 488                          & 0.88        & 1.06            \\
    WPT-Student                    & 12    & 168                          & 0.66        & 0.24            \\
    \bottomrule
    \end{tabular}
}
% \vspace{-2mm}
\label{tab:training}
\end{table}

\begin{table}[ht]
% 控制列间距
\setlength{\tabcolsep}{0.01\linewidth}
\centering
\caption{\textbf{Comparison of planning inference time.}}
% \vspace{-2mm}
\adjustbox{max width=1.0\linewidth}{
    % 在这里放表格
    % hline使用 \toprule \midrule \bottomrule代替，会让竖线出现一个小的间距更好看一些
\begin{tabular}{l|c|cc|c} 
\toprule
\multirow{2}{*}{\textbf{Method}} & \multirow{2}{*}{\textbf{Rwd Model}} & \multicolumn{2}{c|}{\textbf{Planning}} & \textbf{Latency}  \\
                        &                         & Avg. L2 (m)$\downarrow$ & Avg. Col. (\%)$\downarrow$  & (ms)$\downarrow$     \\ 
\midrule
Baseline-T                & -                       & 0.72        & 0.71            & 286      \\
WPT-Teacher                     & -            & 0.62        & 0.14            & 286      \\
WPT-Teacher                     & $\checkmark$                       & 0.61        & 0.11            & 312      \\
\midrule
Baseline              & -                       & 0.88        & 1.06            & 64       \\
WPT-Student                    & -                       & 0.66        & 0.24            & 64       \\
\bottomrule
\end{tabular}
}
\label{tab:inference}
% \vspace{-4mm}
\end{table}

\subsection{Computation and Latency}
We analyze compute and inference cost for the teacher (WPT-Teacher) and the distilled compact student (WPT-Student) in \Cref{tab:training,tab:inference}. Here, {Baseline-T} is the teacher without reward model, and {Baseline} is the student trained without reward model and distillation.

\begin{figure*}[ht]
    \centering
    \includegraphics[width=1\linewidth]{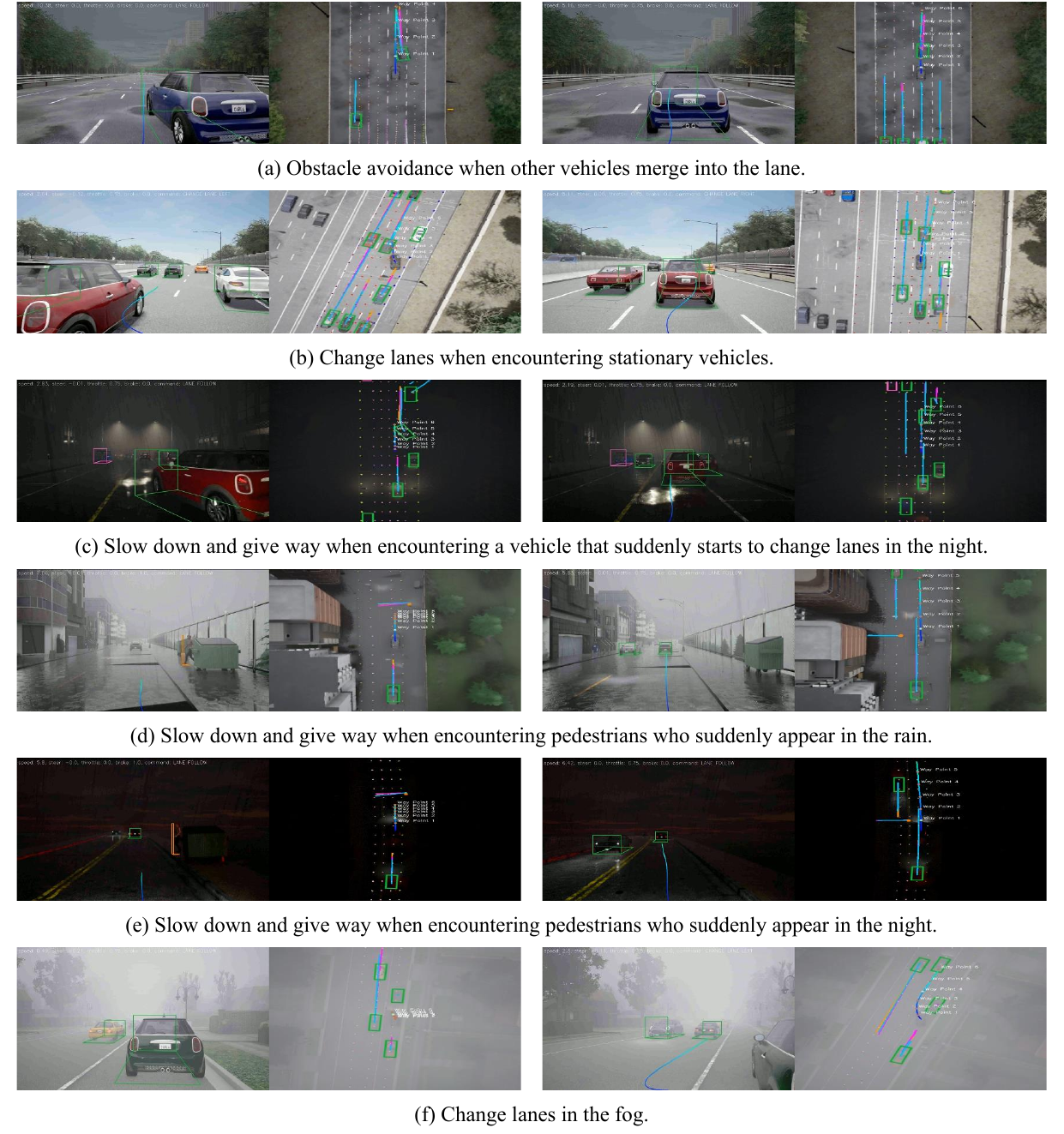}
    % \vspace{-3mm}
    \caption{Visualization of representative closed-loop scenarios from the Bench2Drive benchmark.}
    % \vspace{-4mm}
    \label{fig:bench2drive_vis}
\end{figure*}

\noindent\textbf{Training Process.}
Relative to Baseline-T, WPT-Teacher halves training to 12 epochs and reduces GPU time to 248h (\textbf{–46.6\%}) while improving planning to 0.61m / 0.11\% (\Cref{tab:training}). 
For the lightweight policy, distillation cuts training from 488\,h (Baseline) to 168h (\textbf{–65.6\%}) and improves planning from 0.88m / 1.06\% to 0.66\,m / 0.24\%. 
Thus, WPT-Teacher accelerates convergence and lowers compute for both teacher and student, with especially large relative savings for the compact model.

\noindent\textbf{Inference Latency.}
% At inference (~\Cref{tab:inference}), enabling the reward model online adds a small overhead to the teacher (286$\rightarrow$312\,ms, \textbf{+9\%}) while giving the best safety (0.11\% collisions). Removing the reward model at inference time restores the baseline latency (286\,ms) yet still improves accuracy and safety over the baseline (0.62\,m / 0.14\% vs.\ 0.72\,m / 0.71\%). The distilled student preserves the \textbf{64\,ms} latency of the baseline student but delivers markedly better plans (0.66\,m / 0.24\% vs.\ 0.88\,m / 1.06\%), showing that world-aware distillation transfers most of the benefit without any test-time world-model calls.
On the teacher, enabling the reward model online adds a small overhead (286\,$\rightarrow$\,312ms, but yields the best safety (0.11\% collisions). 
Turning the reward model off restores baseline latency (286ms) while still outperforming Baseline-T (0.62m / 0.14\% vs.\ 0.72m / 0.71\%). 
The distilled student maintains the \textbf{64ms} latency of Baseline yet delivers markedly better plans (0.66m / 0.24\% vs.\ 0.88m / 1.06\%), indicating that world-aware distillation transfers most of the benefit without any test-time world-model calls. 

\subsection{Visualizations}
% 我们提供了Bench2Drive的闭环仿真结果的一些困难场景的可视化结果，如图所示.除此之外，我们还提供了更多的闭环视频数据在补充材料。
We present visualizations of some challenging closed-loop simulation scenarios from the Bench2Drive benchmark, as shown in \cref{fig:bench2drive_vis}. Additionally, more closed-loop visualizations in video format can be found in the supplementary materials.

% Visualization of representative closed-loop scenarios from the Bench2Drive benchmark.

\end{document}